\documentclass{article}

\usepackage{PRIMEarxiv}

\usepackage[utf8]{inputenc} 
\usepackage[T1]{fontenc}    
\usepackage{hyperref}       
\usepackage{url}            
\usepackage{booktabs}       
\usepackage{amsfonts}       
\usepackage{nicefrac}       
\usepackage{microtype}      
\usepackage{lipsum}
\usepackage{fancyhdr}       
\usepackage{graphicx}       
\usepackage{siunitx} 
\graphicspath{{media/}}     

\title{YZS-model: A Predictive Model for Organic Drug Solubility Based on Graph Convolutional Networks and Transformer-Attention.
}

\author{
  Chenxu Wang$^\dag$\\
  Shihezi University \\
  Xinjing, China\\
  zxsdsama@gmail.com,\\
  20221018267@stu.shzu.edu.cn \\
  \And
  Haowei Ming$^\dag$ \\
  Peking University \\
  Beijing, China\\
  \And
  Jian He$^\dag$ \\
  Xinjiang University \\
  Xinjiang, China \\
  \And
  Yao Lu\\
  Shihezi University \\
  Xinjiang, China \\
  \And
  Junhong Chen \\
  South China University of Technology \\
  Guangdong, China \\
  NetEase, Inc. \\
  Zhejiang, China
}

\begin{document}
\maketitle

\let\thefootnote\relax\footnotetext{$\dag$ : Each symbol represented that these authors contributed equally to this work.}

\begin{abstract}
  Accurate prediction of drug molecule solubility is crucial for therapeutic effectiveness and safety. Traditional methods often miss complex molecular structures, leading to inaccuracies. We introduce the YZS-Model, a deep learning framework that innovatively integrates Graph Convolutional Networks (GCN), Transformer architectures, and Long Short-Term Memory (LSTM) networks to enhance prediction precision.GCNs excel at capturing intricate molecular topologies by modeling the relationships between atoms and bonds, providing a detailed understanding of the molecular structure. Transformers, with their self-attention mechanisms, effectively identify long-range dependencies within molecules, capturing global interactions that other models might overlook. LSTMs contribute by processing sequential data, preserving long-term dependencies and integrating temporal information within molecular sequences. This multifaceted approach leverages the strengths of each component, resulting in a model that comprehensively understands and predicts molecular properties.Trained on 9,943 compounds and tested on an anticancer dataset, the YZS-Model achieved an $R^2$ of 0.59 and an RMSE of 0.57, outperforming benchmark models 0.52($R^2$) and 0.61(RMSE). In an independent test, it demonstrated an RMSE of 1.05, improving accuracy by 45.9\%.The integration of these deep learning techniques allows the YZS-Model to automatically learn valuable features from complex data without predefined parameters, efficiently handle large datasets, and adapt to various molecular types. This comprehensive capability significantly improves predictive accuracy and model generalizability. Its precision in solubility predictions can expedite drug development by optimizing candidate selection, reducing costs, and enhancing efficiency. Our research underscores deep learning's transformative potential in pharmaceutical science, particularly for solubility prediction and drug design.
\end{abstract}

\section{Introduction}

In contemporary medical science, the development and application of organic pharmaceuticals hold immeasurable value for improving public health\cite{chin2006drug}. Organic drugs have become critical in combating chronic diseases such as cancer, cardiovascular diseases, and diabetes. For example, some targeted therapies have become essential components of cancer management, significantly improving patient survival rates\cite{soerensen2014improved}. However, despite technological advancements leading to new therapeutic approaches, the path of drug development remains fraught with challenges and high costs. In the United States, the cost of bringing a new drug to market ranges from \$157.3 million to \$1.9508 billion, spanning twelve to fifteen years from initial research to commercial availability\cite{prasad2017research, veljkovic2007application}. This underscores the necessity for efficient and cost-effective methods to streamline the drug development process.

One of the critical aspects of drug development is the accurate prediction of a drug molecule’s water solubility, as it directly influences the drug’s ADME (Absorption, Distribution, Metabolism, and Excretion) properties, ensuring its efficacy and safety\cite{sarisozen2019stimuli}. Accurate solubility predictions can significantly reduce the risk of late-stage failures and optimize the drug design process. Traditional methods for predicting water solubility, such as empirical models (e.g., Jouyban-Acree model), quantitative structure-activity relationship (QSAR) models, and physicochemical computational methods, although effective in some cases, often perform poorly when applied to molecules with complex structures\cite{taskinen2000prediction, jouyban2006silico, duchowicz2008new}. These methods have the following main limitations: restricted predictive accuracy, especially for molecules that are structurally complex or have novel characteristics\cite{jorgensen2002prediction}; a heavy reliance on extensive experimental data, limiting their effectiveness in cases with few samples or novel molecular structures\cite{lipinski1997experimental}; the requirement for expensive computational resources, particularly in physicochemical approaches\cite{dadmand2019solubility}; and insufficient comprehension of intricate molecular interactions, hindering a comprehensive reflection of molecular solubility behavior\cite{Christopher2001Experimental}. These challenges have prompted the scientific community to explore more advanced and precise prediction methods.

Within the current pharmaceutical research landscape, deep learning provides a novel approach that significantly improves the precision and efficiency of solubility predictions. Deep learning offers distinct potential advantages over traditional methodologies: automatic feature learning, where deep learning models can inherently learn and extract valuable features from complex data without relying on predefined physicochemical parameters, enabling the handling of highly complex molecular structures\cite{cui2020improved}; the capacity to handle large datasets, as deep learning can adeptly manage and analyze extensive datasets, including those derived from high-throughput experiments, enhancing the predictive model's generalizability and accuracy\cite{ghahremanpour2023ensemble}; adaptability and flexibility of models, allowing deep learning models to be tailored to predict particular molecular types by modifying network structures and parameters, such as employing various neural network types like CNNs and RNNs to better process spatial and temporal molecular data\cite{gao2020accurate}; and swift iteration and updating, enabling deep learning models to rapidly integrate new experimental data and align with emerging research and market needs, thereby expediting the drug development timeline\cite{liu2023fragment}. Therefore, deep learning is chosen as the primary predictive model due to its outstanding performance and strong adaptability to unknown or complex data, significantly expediting the drug development process and aiding in the early identification of candidates with high therapeutic potential, thus greatly reducing both cost and time in development.

Numerous studies have demonstrated the significant role of deep learning in the fields of drug discovery and materials science. For example, Mater et al. discussed the applications of deep learning in drug and material design and synthesis planning, highlighting its potential to enhance and expedite the drug development workflow\cite{mater2019deep}. Aliper et al. showed how deep neural networks can categorize drugs using transcriptome data, which helps predict drug effectiveness and side effects more accurately\cite{aliper2016deep}. Additionally, Walters et al. applied deep learning to predict drug molecule properties, providing essential insights during the initial phases of drug design to circumvent expensive modifications during later development stages\cite{walters2020applications}.

Deep learning has also demonstrated its importance and effectiveness in predicting the solubility of small molecule drugs. For instance, Alessandro employed a novel approach using recurrent neural networks to tackle the issue of cyclic structural information loss\cite{lusci2013deep}. Recurrent neural networks not only address this issue but also enable the automatic learning of appropriate molecular representations, reducing the need for extensive domain expertise. Beyond simple RNNs, Waqar added attention mechanisms to convolutional neural networks, facilitating the broad and automatic acquisition of solubility-relevant chemical features\cite{ahmad2023attention}. In addition to innovations in neural network architecture, Tong suggested using molecular descriptors instead of structural forms for training neural networks\cite{deng2020prediction}. This approach not only simplifies the representation of complex molecular structures but also improves training efficiency. Moreover, researchers such as Sumin have made innovative improvements to molecular inputs, with the MFPCP method incorporating molecular fingerprints and physicochemical properties into a combined descriptor for high-performance solubility prediction\cite{lee2022novel}. These studies not only emphasize the impact of deep learning in pharmaceutical research but also demonstrate its effectiveness in addressing complex drug design challenges.

To address the unresolved challenges in predicting drug solubility, our research innovatively integrates advanced deep learning models to develop the YZS-Model. Traditional methods for predicting water solubility—such as empirical models, QSAR models, and physicochemical computational methods—have shown limited success, particularly with structurally complex or novel molecules. These methods often suffer from restricted predictive accuracy, a heavy reliance on extensive experimental data, high computational costs, and an inadequate understanding of intricate molecular interactions. The YZS-Model addresses these critical issues by combining Graph Convolutional Networks\cite{duvenaud2015convolutional}  (GCNs), Long Short-Term Memory networks\cite{hochreiter1997long} (LSTM), and Transformer\cite{vaswani2017attention} architecture to analyze SMILES strings and predict compound solubility. This innovative model leverages multiple deep learning technologies to capture both the global graph structure and sequential information of molecules, representing a novel approach to drug solubility prediction.

The methodology begins with GCNs to capture the holistic graph structure of molecules, enabling a detailed understanding of their complex interconnections. Following this, LSTM and Transformer networks handle the sequential data within molecules. The Transformer's self-attention mechanism is particularly effective at recognizing long-distance molecular dependencies, while the LSTM processes and integrates molecular relationships not entirely captured by the Transformer. This multifaceted approach significantly enhances the model’s ability to understand and predict molecular properties.Specifically, the GCN component allows for a comprehensive analysis of molecular topology, capturing key structural features. The Transformer's robust self-attention mechanism efficiently identifies and processes long-range dependencies within molecules\cite{vaswani2017attention}. Meanwhile, the LSTM ensures that the model fully understands and integrates complex sequence information. This multi-technique methodology provides a fresh perspective for predicting drug solubility, demonstrating the vast potential of deep learning in chemical data analysis and pharmaceutical development.

Through these innovations, the YZS-Model not only addresses the limitations of traditional methods but also significantly reduces the risk of late-stage drug development failures. Accurate solubility predictions can optimize the drug design process, leading to more effective and safer drugs. Therefore, this research is essential for advancing pharmaceutical sciences, improving public health outcomes, and reducing the overall cost and time associated with bringing new drugs to market.

Beyond its innovativeness, the model also performs exceptionally in benchmark tests. It achieved a coefficient of determination ($R^2$) of 0.55 and a root mean square error (RMSE) of 0.59 on a dataset of 62 anticancer organic drugs, clearly surpassing the existing benchmark model, AttentionFP, which scored 0.52 and 0.61 respectively. Furthermore, on a second test set compiled by Linas et al., it reached an RMSE of 1.05, significantly outperforming the benchmark model's 1.28. This breakthrough not only validates our model's superiority but also its high accuracy and generalization capability in practical applications.

Regarding the model's interpretability, we conducted a two-fold feature importance analysis. The first aspect involves a quantitative ranking of the importance of features used by our model, identifying potential correlations between specific molecular features and solubility. The second aspect involves a qualitative chemical analysis of features and elements likely to have significant correlations, using the number and strength of hydrogen bonds, as well as the number of hydrophilic and hydrophobic groups and specific chemical molecules as examples. By deeply analyzing each feature and elaborating on the model's predictive logic, we provide a solid theoretical and empirical foundation for practical applications. This assists researchers in effectively predicting and optimizing the physicochemical properties of drugs in the early stages of pharmaceutical development, thereby reducing the risk of development failures and enhancing efficiency.

In summary, this study not only showcases the immense potential of deep learning for predicting the solubility of drug molecules but also offers a new solution for reducing drug development cycles and costs\cite{liu2023comprehensive}. As data-driven research methodologies become increasingly prevalent in drug development, we believe this marks the dawn of a new era in pharmaceutical research. We look forward to further breakthroughs in this field that could significantly contribute to human health.

The main contributions of this paper are as follows:

\textbf{Multi-model Integration and Interaction Design}: The core innovation of the YZS-Model lies in its unique multi-model integration and interaction design, combining GCNs, LSTM networks, and Transformer architecture. This design enables the model to comprehensively capture both local and global information of molecules, significantly enhancing the accuracy and robustness of predicting organic drug solubility. By integrating multiple advanced deep learning models, the YZS-Model successfully achieves a multi-level and multi-perspective analysis of complex molecular structures.

\textbf{Construction of Molecular Structure Feature Sequences}: This model iteratively extracts features for each atom within the molecule over k iterations, obtaining the features within each molecule's k-neighborhood. These final molecular structure feature sequences are used as inputs to the Transformer, providing a solid foundation for subsequent solubility predictions. This process effectively captures complex interactions and long-range dependencies within molecules, thereby enhancing the model's predictive accuracy.

\textbf{Sequence Dependency Analysis with LSTM}: In the YZS-Model, LSTM is used to handle the sequential data within molecules, particularly capturing long-term dependencies in molecular sequences. With its unique memory mechanism, LSTM addresses the gradient vanishing problem in traditional RNNs when processing long sequences. By integrating various aspects of molecular relationships, LSTM can further uncover temporal dependencies within molecular structure sequences, enhancing the model's predictive capability. This allows the YZS-Model to excel in handling complex molecular sequence data, enabling more accurate predictions of drug solubility.

\section{Materials and Methods}

\subsection{Training and Testing}

This research was conducted using a dataset gathered by Cui, Q., and colleagues, which included 9,944 organic compounds\cite{cui2020improved}. The dataset is structured with three primary components for each entry: the Simplified Molecular Input Line Entry System (SMILES), International Chemical Identifier (InChIKeys), and solubility (log S). SMILES represents molecular structures using ASCII strings, which not only streamline the electronic documentation of compounds but also enhance their computability and analytical processing. Log S, the logarithmic measure of molecular solubility in mol/L, is determined under normal conditions at room temperature, atmospheric pressure, and a pH of 7.0, with values spanning from -18.22 to 1.70.

\begin{figure}[h]
  \centering
  \includegraphics[width=0.6\linewidth]{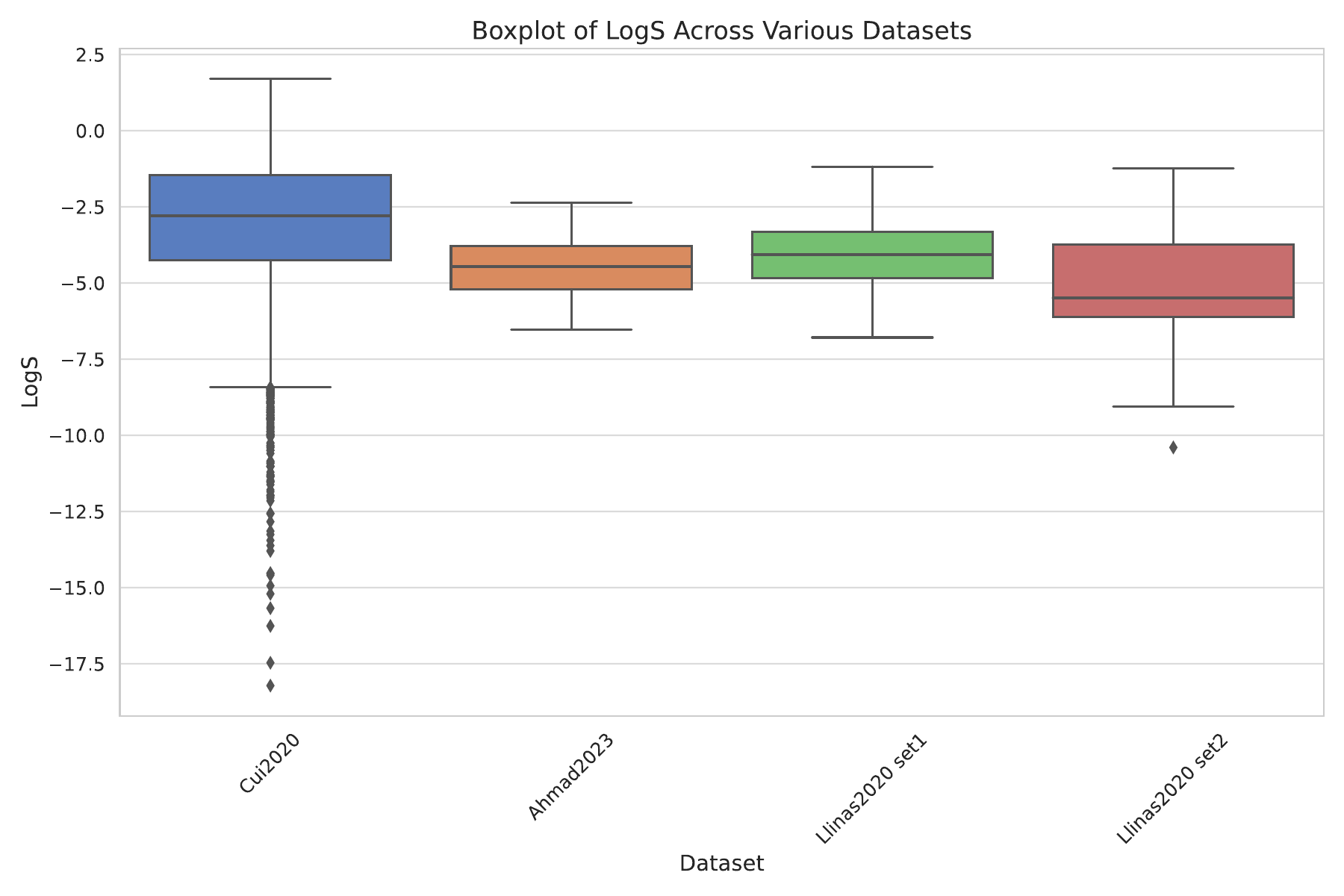}
  \caption{Solubility (log S) Distribution Across Datasets.}
  \label{fig:fig1}
\end{figure}

To assess the effectiveness of our model, three separate test sets were employed. The initial test set, curated by Waqar Ahmad et al., contained 62 clinically validated anticancer compounds, showcasing log S values between -6.52 and -2.36, with an average of -4.48 and a median of -4.46\cite{ahmad2023attention}. This demonstrates the diversity and experimental relevance of the dataset used. Further, we used two additional datasets released by Llinas2020 for baseline comparisons—one with 100 entries and another with 32—to thoroughly evaluate the model's performance\cite{llinas2020findings}. The distribution of log S values in other datasets is depicted in the accompanying figure \ref{fig:fig1}.

In terms of data processing, all datasets were standardized prior to model training and testing. This included steps such as removing duplicates, shuffling the molecules, and feature selection. Specifically, SMILES and InChIKeys were matched one-to-one. We crawled InChIKeys information from the ACS database and compared it with SMILES to prevent mismatches. In the ablation experiments, due to the involvement of training set ablation, to prevent the concentration of any specific feature, we reshuffled the data based on a normal distribution. Additionally, the log S data, as collected by the original compilers, had already undergone normalization to avoid numerical discrepancies that could impact model training.

\subsection{Data Preprocessing}

In our study, we employed the RDKit library to convert SMILES (Simplified Molecular Input Line Entry System) strings into molecular objects.Influenced by Ahmad's paper\cite{ahmad2023attention},these molecular objects serve as the foundational data structure for further extraction and analysis of molecular features. RDKit is a widely used cheminformatics toolkit that allows for the processing and manipulation of chemical information, and it was essential for our feature extraction pipeline. 

To accurately represent the intricate details of each molecule, we constructed a feature tensor that stores comprehensive information about the core attributes of each molecule. These attributes include:

Type of Atoms: The elements present in the molecule, such as carbon (C), nitrogen (N), oxygen (O), etc. Each element is encoded as a binary feature where 0 indicates the absence and 1 indicates the presence of that particular element. For a more detailed representation, each atom type is represented in a 66-bit one-hot encoded vector, as outlined in Table \ref{tbl:atom_features}.

\textbf{Number of Bonds}: The number of bonds between atoms, indicating the molecule's connectivity. This feature captures the total number of bonds each atom forms, represented in an 8-bit binary encoding.

\textbf{Formal Charges}: The formal charge on each atom, represented as a single integer value indicating the charge.

\textbf{Free Radical Electrons}: The number of unpaired electrons on each atom, represented as a single integer.

\textbf{Hybridization Types}: The hybridization state of each atom, such as sp, sp$^2$, sp$^3$, etc. This is encoded using a 7-bit one-hot encoding.

\textbf{Aromaticity}: Whether the atom is part of an aromatic system, encoded as a binary feature (true/false).

\textbf{Number of Directly Bonded Hydrogen Atoms}: This feature indicates the number of hydrogen atoms directly bonded to each atom, encoded using a 5-bit one-hot encoding.

\textbf{Chirality and Stereochemistry}: Information about the chirality (whether the atom is chiral) and the type of chirality (R or S configuration), with chirality being a binary feature and chirality type being a 2-bit one-hot encoding.

The complete set of atomic features utilized in our molecular object analysis is detailed in Table \ref{tbl:atom_features}.

\begin{table}[h]
\small
\centering
  \renewcommand{\arraystretch}{1.3}
  \caption{\ Detailed Attributes of Atomic Features Utilized in Molecular Object Analysis}
  \label{tbl:atom_features}
  \begin{tabular*}{0.80\textwidth}{@{\extracolsep{\fill}}lll}
    \hline
    Atom Features & Size & Details \\
    \hline
    Symbol & 66 & [K, Y, V, Sm, Dy, In, Lu, Hg, Co, Mg, Cu, Rh,\\
     &  &  Hf, O, As, Ge, Au, Mo, Br, Ce, Zr, Ag, Ba, N,\\
     &  &  Cr, Sr, Fe, Gd, I, Al, B, Se, Pr, Te, Cd, Pd, Si,\\
     &  &  Zn, Pb, Sn,Cl, Mn, Cs, Na, S, Ti, Ni, Ru, Ca,\\
     &  &  Nd, W, H, Li, Sb, Bi, La, Pt, Nb, P, F, C, Re,\\
     &  &  Ta, Ir, Be, Tl]\\
    Degree & 8 & Total direct connected atoms.\\
    Formal Charge & 1 & Total formal charges.\\
    Electrons & 1 & Total radical electrons.\\
    Hybridization & 7 & s,sp,sp$^2$,sp$^3$,sp$^3$d,sp$^3$d$^2$,other.\\
    Aromaticity & 1 & Atom is in aromatic system (true/false).\\
    Hydrogens & 5 & Total direct connected hydrogens.\\
    Chirality & 1 & Atom is chiral (true/false).\\
    Chirality type & 2 & R or S.\\
    \hline
  \end{tabular*}
\end{table}

For capturing the internal bond information within the molecules, we used two key data structures: `edge\_attr' and `edge\_index'.

\textbf{`edge\_attr'}: This stores attributes related to the bonds, such as the type of bond (single, double, triple, aromatic), whether the bond is conjugated, its aromaticity, and whether the bond is part of a ring. Each bond type is encoded in a 4-bit binary feature as shown in Table \ref{tbl:bond_features}.

\textbf{`edge\_index'}: This records the connectivity between atoms, including atom indices and types of bonds, providing a detailed structural framework of the molecule.
The bond features are summarized in Table \ref{tbl:bond_features}.

Regarding internal bond information of the molecules, we utilized two key data structures: `edge\_attr' and `edge\_index'. `edge\_attr' primarily stores attributes related to the bonds, such as the type of bond (single, double, triple), whether the bond is conjugated, aromaticity, and whether the bond is part of a ring. `edge\_index' records the connectivity between atoms, including atom indices, type of bonds, and other relevant information, defining the structural framework of the molecule, as detailed in Table \ref{tbl:bond_features}.

\begin{table}[h]
\small
\centering
  \renewcommand{\arraystretch}{1.3}
  \caption{\ Detailed Attributes of Bond Features Utilized in Molecular Object Analysis}
  \label{tbl:bond_features}
  \begin{tabular*}{0.8\textwidth}{@{\extracolsep{\fill}}lll}
    \hline
    Bond Features & Size & Details \\
    \hline
    Type & 4 & Single, double, triple, aromatic.\\
    Conjugation & 1 & Bond is conjugate (true/false).\\
    Ring & 1 & Bond is in ring (true/false).\\
    Stereo & 4 & StereoNone, StereoAny, StereoZ, StereoE.\\
    \hline
  \end{tabular*}
\end{table}

These detailed features are subsequently integrated into a `torch\_geometric' Data object, which includes the node features, edge connectivity, and their attributes, forming a complete graph data structure for analysis and model training.

\begin{figure}[!h]
  \centering
  \includegraphics[width=1\linewidth]{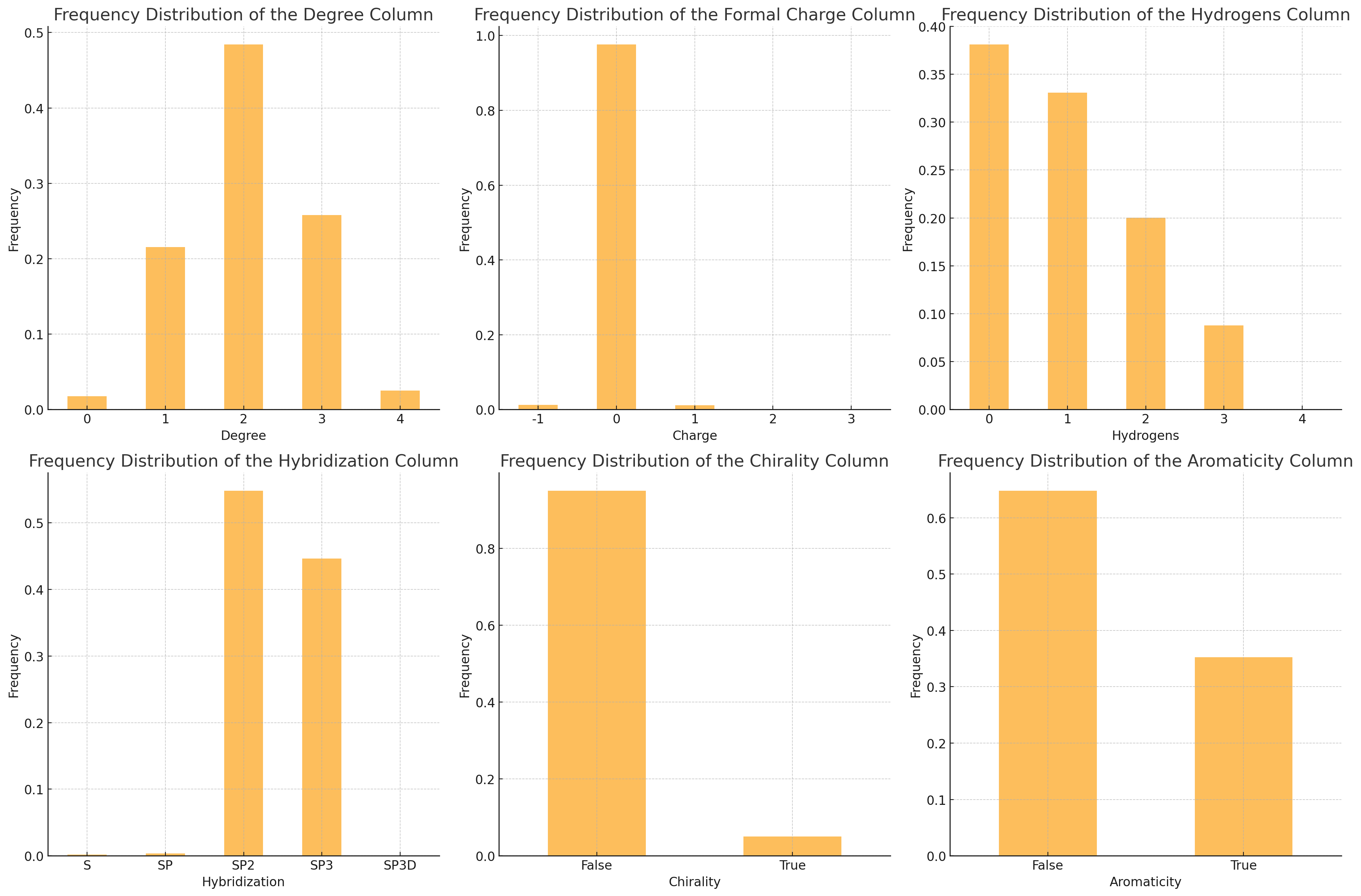}
  \caption{Distribution of Molecular Features in the Training Dataset.}
  \label{fig:figx}
\end{figure}

To visualize the distribution of these molecular features, we will present the figure \ref{fig:figx} that depicts the frequency and diversity of the key attributes within our dataset. This figure will help illustrate the richness of the data and the broad range of molecular characteristics captured by our preprocessing pipeline.

This visualization provides insight into the variability and complexity of the molecular structures analyzed in our study, highlighting the diversity of atomic and bond features present in the dataset.

\subsection{10-Fold Data Split}
In our research, we adopted the version of the Cui2020 dataset segmented by Ahmad et al. for the training and testing phases of our machine learning model\cite{ahmad2023attention, cui2020improved}. The team led by Ahmad et al. randomly split the original Cui2020 dataset into ten equal and independent subsets, each termed a ``fold''. To evaluate our model's efficacy, we implemented a 10-fold cross-validation technique: during each cycle, the model trained on nine folds and was tested on the remaining one. This procedure was replicated ten times, ensuring that each fold was used as a test set once.

This method of cross-validation enables the assessment of the model under various training-test data configurations, thereby offering a robust and credible estimate of its performance. We documented the model's test set performance in all ten iterations, selecting the optimal model configuration for additional training and testing. The objective of this phase was to pinpoint a model configuration that provides the strongest generalization capability across the complete dataset.

\subsection{Molecular Feature Extraction}

In this study's model, the feature extraction step is crucial. Influenced by Ahmad's paper\cite{ahmad2023attention} on feature extraction, we utilized the RDKit tool to convert the molecular SMILES representation into Mol objects. We then traversed all atoms and chemical bonds, using RDKit's member methods to accurately extract various features of atoms and bonds. These features were organized into tables (see Table \ref{tbl:atom_features} and Table \ref{tbl:bond_features}). On this basis, atoms and bonds were abstracted as nodes and edges in graph theory, and these graph features were converted into a format suitable for deep learning using the PyG library, with edges and vertices represented by their respective high-dimensional feature vectors. The node feature vectors have a dimension of 92, while the edge feature vectors have a dimension of 10.

We extracted the majority of features present in organic compounds to serve as model features. Since these feature representations cannot be directly processed by the model, we employed one-hot encoding to encode all multi-type variables, converting them into binary forms of zero or one. Specifically, element symbols are encoded using 66-bit one-hot vectors, which cover the vast majority of atoms that constitute organic compounds. Additionally, hybridization features are encoded using 7-bit one-hot vectors, encompassing the most common hybridization forms found in organic materials. This encoding strategy not only simplifies the processing of categorical variables by machine learning algorithms but also ensures the independence of features and effectively prevents model bias caused by changes in feature order \cite{cerda2018similarity}, thereby streamlining the model's feature extraction and utilization process.

\subsection{Graph Neural Network}

Graph Neural Networks (GNNs) represent a cutting-edge deep learning framework, devised for processing and predicting data that takes the form of graphs. In contrast to conventional deep learning models, GNNs adeptly manage datasets with complex interrelations and varying dimensions, thus exhibiting exceptional utility in numerous domains. GNNs function by precisely modeling the interactions among nodes and edges in a graph, thereby learning and extracting profound characteristics of graph data. In this framework, each node signifies an entity, while edges represent the relationships among these entities. The unique aspect of GNNs lies in their iterative process of updating node states, which enhances node representations by leveraging information from adjacent nodes. The input for a GNN consists of an undirected graph $G$.

\begin{equation}
    G = (V, E)
\end{equation}

Here, $V$ denotes the set of all nodes within the graph, and $E$ refers to the collection of edges that directly connect the nodes. Each node $v\in V$ is associated with an initial feature vector $x_v$, encoding both node and edge characteristics. The objective of a GNN is to devise a function $f$ that produces a sophisticated feature representation $h_v$ for every node, capable of encapsulating both the local structural dynamics and the intricate interactions of node features.

In the case of node $v$ , updating its hidden state each round consists of the following steps:
\begin{enumerate}
    \item \textbf{Message Passing}: First, compute the message passed to node $v$ from each of its neighboring nodes $u \in N(v)$ using the equation:
    \begin{equation}
        m_{uv}^{(t+1)}=M(h_v^{(t)},h_{u}^{(t)},x_u,x_v,e_{uv})
    \end{equation}
    where $M$ stands for the message function, $h^{(t)}_v$ and $h^{(t)}_u$ are the hidden states of nodes $v$ and $u$ in iteration $t$, $x_v$ and $x_u$ are the nodes' initial features, and $e_{uv}$ is the feature of the edge connecting $v$ and $u$.
    
    \item \textbf{Aggregation}: Node $v$ aggregates the messages received from all its neighbors into a single composite message using:
    \begin{equation}
        a_v^{(t+1)}=A(\{m_{uv}^{(t+1)}:u\in N(v)\})
    \end{equation}
    $A$ represents the aggregation function, which may be summing, averaging, or taking the maximum value.

    \item \textbf{State Update}: Finally, the hidden state of node $v$ is updated with the update function:
    \begin{equation}
        h^{(t+1)}_v=U(h_v^{(t)},a_v^{(t+1)},x_v))
    \end{equation}
    $U$ denotes the update function, which could be a straightforward linear transformation with a nonlinear activation or a more intricate neural network.
\end{enumerate}
These procedures allow GNNs to iteratively update the hidden states of each node in the graph. Eventually, the hidden states converge, encapsulating and reflecting the features of their neighboring nodes and the graph's structural properties. This iterative cycle is generally repeated for several rounds until the nodes' hidden states stabilize or reach the pre-set number of iterations.

The output of a GNN is tailored to fit specific tasks such as node classification, graph classification, or link prediction. In the case of node classification, the network maps the final hidden state $h^{(T)}_v$ of a node to its predicted label using a readout layer $R$:
\begin{equation}
    \hat y_v=R(h^{(T)}_v)
\end{equation}
where $T$ denotes the total number of iteration rounds completed, and $\hat y_v$ represents the predicted label for node $v$. This approach allows the GNN to deduce the label of each node by integrating information from its features and neighborhood over multiple iterations.

\subsection{Graph Convolution Network}

Based on a comparison of various existing GNN model architectures, we introduced one of the classic applications of GNN, the Graph Convolutional Network (GCN) architecture, to process graph data. GCN utilizes convolution operations on graph structures to gradually and effectively capture the feature relationships between nodes and their neighbors. In this study, by applying the GCN architecture, we can fully capture the molecular features within the k-radius neighborhood of each atom. This approach fully leverages the information within the graph structure of organic molecules, providing robust support for the prediction and analysis of molecular properties.\cite{feng2021cross}.

\subsection{Self-Attention Transformer}

In our research, we utilized a Self-Attention Transformer architecture designed to analyze and predict the solubility of organic drugs. This architecture differs from conventional deep learning models such as RNNs and LSTMs by employing a self-attention mechanism that captures global dependencies within sequences. This feature significantly improves the model's capability to process long-distance dependencies\cite{xu2019graph}.

The self-attention mechanism enables the model to focus on any position within the sequence, calculating representations for each position and dynamically adjusting its focus to detect long-range dependencies. For an input sequence representation $X\in \mathbb{R}^{n\times d}$ where $n$ is the length of the sequence and $d$ is the dimension of features, the self-attention layer computes attention scores for each element in relation to others, expressed as:
\begin{equation}
    \mathrm{Attention}(Q,K,V)=\mathrm{softmax}\left(\frac{QK'}{\sqrt{d_k}}\right)V
\end{equation}
In this formula, $Q$, $K$, and $V$ denote the query, key, and value matrices, respectively, derived by multiplying the input sequence $X$ with corresponding weight matrices; $d_ k$ stands for the dimension of the key vectors.

\begin{figure}[h]
  \centering
  \includegraphics[width=0.45\linewidth]{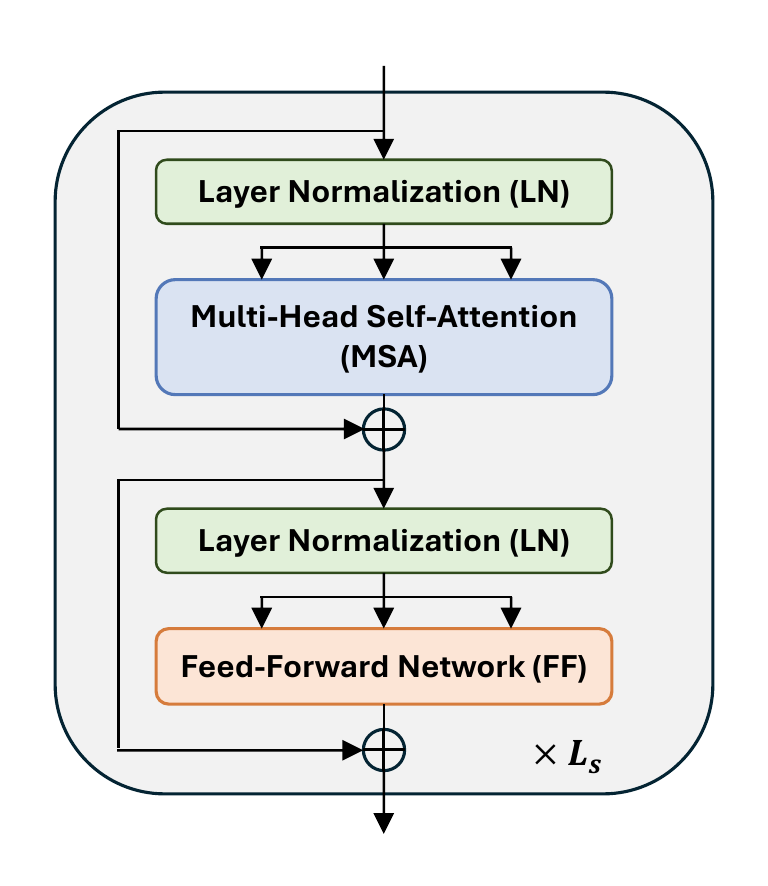}
  \caption{An overview of the Transformer model structure, LN, MSA, and FF, demonstrates critical steps in sequence data processing.}
  \label{fig:fig2}
\end{figure}

The Transformer architecture is built from encoder and decoder components, consisting of several repeating layers. Each layer includes a multi-head self-attention mechanism and a position-wise fully connected feed-forward network, both supplemented by residual connections and layer normalization, as illustrated in Figure \ref{fig:fig2}.
\begin{enumerate}
    \item \textbf{Multi-Head Self-Attention}: This component comprises parallel-executed self-attention layers, enabling the model to capture diverse information across multiple subspaces, thereby enhancing its capability to represent various features.

    \item \textbf{Feed-Forward Network}: Each position's feature vector undergoes two linear transformations and a nonlinear activation within this network, promoting further information synthesis.
\end{enumerate}

In the context of predicting drug solubility, the Transformer leverages feature vectors refined by graph convolutional networks to understand the interactions among atoms and functional groups within a molecule's neighborhood.

\subsection{Long Short-term Memory}
In our research, Long Short-Term Memory (LSTM) networks have been utilized to enhance the analysis and prediction of organic drug solubility. LSTMs, advanced variants of Recurrent Neural Networks designed to capture long-term dependencies in sequence data, are particularly tailored in this study to optimize outputs derived from Self-Attention Transformer architectures, revealing deeper sequential dependencies in molecular sequences.

LSTMs effectively solve the gradient vanishing or exploding problems seen in traditional RNNs when handling long sequences, thanks to their distinctive gating mechanism. This includes:
\begin{enumerate}
    \item \textbf{Forget Gate}: It assesses which information should be removed from the cell state, based on the current input and the previous hidden state.

    \item \textbf{Input Gate}: It decides which new information is to be incorporated into the cell state.

    \item \textbf{Output Gate}: It determines what the next hidden state should be, computed from the current cell state influenced by the gates' outputs.
\end{enumerate}

Thus, by handling the outputs from the Self-Attention Transformer, the LSTM not only bolsters the model's grasp of sequence dependencies but also robustly supports precise predictions of solubility in organic drugs. This method allows us to exploit complex interactions within drug molecular structure sequences, providing more detailed and accurate feature representations for predicting solubility.

\subsection{YZS-Model}
\begin{figure}[h]
  \centering
  \includegraphics[width=1\linewidth]{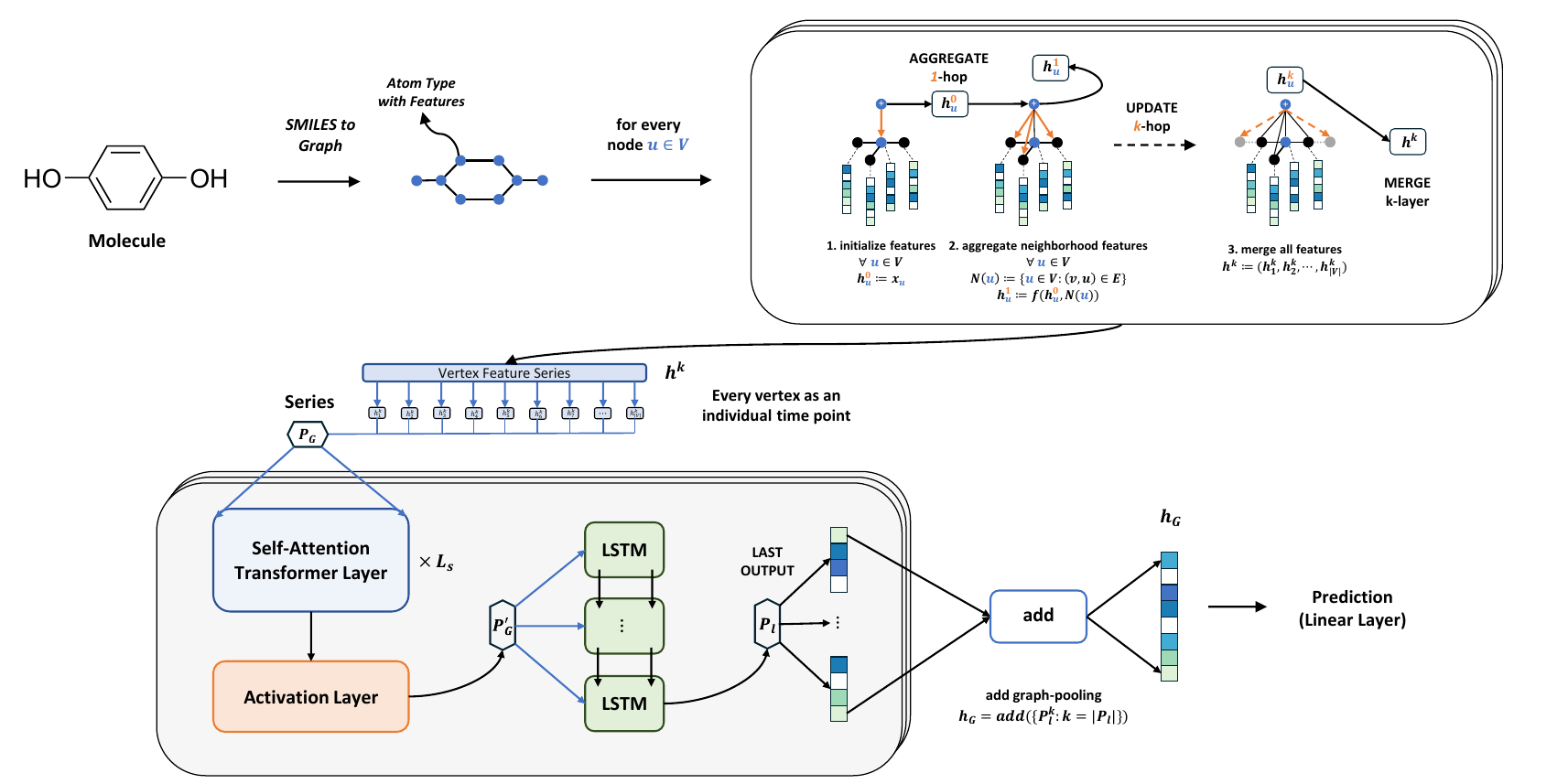}
  \caption{Architecture of the YZS-Model. Initially, drug molecules are converted from SMILES notation into a graph representation, with each atom encoded as a node and bonds as edges. Features are aggregated through a Graph Convolutional Network (GCN) to capture structural information. Subsequently, these features are processed by a self-attention Transformer layer to capture global dependencies. The sequence dynamics are then analyzed via an LSTM layer. Finally, graph pooling and Linear Layer are used to output solubility predictions.}
  \label{fig:fig3}
\end{figure}

In this study, we introduced the YZS-Model, which aims to precisely predict the solubility of organic drugs. This model combines Graph Convolutional Networks (GCN), Transformer architecture, and Long Short-Term Memory networks (LSTM) to thoroughly investigate the spatial structure and sequence dynamics of drug molecules, as illustrated in Figure \ref{fig:fig3}.

After converting SMILES into graph representations, the nodes of the graph correspond to the feature vectors of each atom, while the edges represent the bonds. Therefore, the first step of the model involves processing the graphical representation of drug molecules using a Graph Convolutional Network (GCN), which effectively reveals the spatial structural features of the molecules. By analyzing the interactions between atoms and their chemical bonds, the GCN layer abstracts information that describes the molecular topology. It outputs aggregated feature vectors for each atom within the k-radius neighborhood, which are then combined into a feature sequence and passed to the next module. This approach not only captures the overall molecular information but also considers the influence of neighboring atoms.

Within a molecule, atoms can interact with each other even if they are relatively distant, potentially affecting solubility. Therefore, a Transformer architecture is used to analyze molecular features from a global perspective. Based on the feature sequence provided by the previous module, the Transformer leverages the self-attention mechanism to identify extensive dependencies between atoms within the molecule, regardless of their spatial distance in the molecular structure. This stage of analysis focuses on the intricate details of the molecular microstructure and captures the comprehensive internal interactions, resulting in an advanced feature representation that encompasses extensive global information. The module ultimately outputs a tensor sequence containing the attention-distributed molecular structure sequence and features, which is then handed over to the LSTM for further processing.

Subsequently, the LSTM layer is employed to deepen the understanding of the features generated by the Transformer architecture. Specifically, some features produced by the Transformer may have little or no impact on the solubility of organic compounds. The LSTM layer selectively forgets these less impactful features, retaining only the more significant ones, thus enhancing the model's ability to extract important information. Additionally, considering that the molecular structure sequence output by the Transformer might contain temporal dependencies, and the Transformer's mechanism tends to be less effective in capturing such dependencies, the LSTM’s unique gated mechanism effectively captures the temporal dependencies within the sequence data. The output sequence from the final LSTM layer is then passed to the next stage. This processing phase is crucial for revealing the dynamic characteristics of molecular features as they change over the sequence, providing deep temporal insights for solubility prediction.

Finally, a fully connected layer translates the output of the LSTM layer into the final solubility prediction value. This layer not only integrates the high-dimensional features from the GCN, Transformer, and LSTM but also projects these features linearly to relate them to the solubility of drug molecules. The design of the entire model aims to achieve an effective mapping from the spatial and temporal characteristics of the molecules to their physicochemical property of solubility, ensuring high accuracy and reliability in predicting the solubility of organic drugs.

\subsection{Implementation Details}

In our research, we focused on optimizing the hyperparameters of the YZS-Model to improve its predictive accuracy for the solubility of organic drugs. Using the hyperopt library and the Tree-structured Parzen Estimator (TPE) algorithm, we conducted a systematic exploration and evaluation of several critical hyperparameters, such as learning rate, dimensions of feature vectors, Dropout rate, depth of Transformer layers, number of heads in the multi-head self-attention, and batch size, to discover the best model configuration\cite{bergstra2011algorithms}. The hyperopt training was configured to stop after 200 iterations, with the details documented in a log file, as illustrated in the provided Table \ref{tbl:table2}.

\begin{table}[h]
\small
\centering
  \renewcommand{\arraystretch}{1.3}
  \caption{\ Detailed YZS-model hyperopt training parameters}
  \label{tbl:table2}
  \begin{tabular*}{0.6\textwidth}{@{\extracolsep{\fill}}lll}
    \hline
    Item & Range & Selection method\\
    \hline
    lr & (0.0003,0.0007) & hp.uniform\\
    dim & (92,128,2) & hp.quniform\\
    dropout & (0.25,0.35) & hp.uniform\\
    depth & [2,4,6,8,12] & hp.choice\\
    heads & [4,8,12,16] & hp.choice\\
    batch\_size & (24,72,8) & hp.quniform\\
    \hline
  \end{tabular*}
\end{table}

We ultimately set the depth of the Transformer layers to six, with each layer having eight heads with a dimension of 92, and an Linear Layer dimension of 256, while setting the Dropout rate to 0.25. For the YZS-Model, the number of input features was adjusted to 92, with the feature dimensions increased to 128 and the Dropout rate finely adjusted to 0.2519. These precise modifications, underpinned by rigorous experimental and validation efforts, markedly enhanced the YZS-Model's performance in predicting solubility.

\subsection{Evaluation Metrics}

In our research, the model's performance is assessed using two key metrics: the coefficient of determination ($R^2$) and the Root Mean Square Error (RMSE).

\textbf{Coefficient of Determination ($R^2$)}: This metric measures the goodness of fit of the model to the observed data. It reflects the model's ability to explain the variability of the dependent variable, with values ranging from 0 to 1. A higher $R^2$ value indicates a better fit. It is computed as follows:
\begin{equation}
    R^2 = 1 - \frac{\sum_{i=1}^{n}(y_i - \hat{y}_i)^2}{\sum_{i=1}^{n}(y_i - \bar{y})^2} 
\end{equation}
Where $y_i$ is the actual value, $\hat{y}_i$ is the model's prediction, $\bar{y}$ is the average of the actual values, and $n$ is the sample size.

Root Mean Square Error (RMSE): RMSE quantifies the model's prediction accuracy by calculating the square root of the average of the squared discrepancies between the actual and predicted values. A smaller RMSE value indicates greater accuracy. Its formula is:
\begin{equation}
    \mathrm{RMSE} = \sqrt{\frac{1}{n} \sum_{i=1}^{n}(y_i - \hat{y}_i)^2} 
\end{equation}
Both metrics offer a method to quantitatively assess and delve into the model's efficacy in predicting the solubility of organic drugs.

\section{Results and Discussion}

\subsection{Performance of YZS-Model}

In our research, two metrics were employed to thoroughly evaluate the performance of the proposed model: the coefficient of determination ($R^2$) and the Root Mean Square Error (RMSE). The $R^2$ index measures the model's ability to explain data variability, whereas RMSE assesses the deviations between predicted and actual values. The YZS-Model was evaluated using a ten-fold version of Cui, Q.'s dataset partitioned by Ahmad et al., with Ahmad's AttentiveFP model used as a comparative baseline. Training results shown in Table 3 reveal that our YZS-Model surpasses the baseline model, reaching peak performance.

Further validation was conducted using a dataset of 62 anticancer compounds compiled by Ahmad, as displayed in Table \ref{tbl:AFPans}. The outcomes indicate that the YZS-Model attained scores of 0.59 in RMSE and 0.55 in $R^2$, substantially outperforming the baseline AttentiveFP model's scores of 0.61 and 0.52 and other traditional models, underscoring our model's superior performance.

\begin{table}[h]
\small
\centering
  \renewcommand{\arraystretch}{1.4}
  \caption{\ AttentiveFP Dataset Results Comparison}
  \label{tbl:AFPans}
  \begin{tabular*}{0.60\textwidth}{@{\extracolsep{\fill}}lll}
    \hline
    Model & R$^2$ & RMSE \\
    \hline
    GIN-based GNN & 0.21 & 0.78 \\
    SGConv-based GNN & 0.09 & 1.98 \\
    GAT-based & 0.38 & 0.69 \\
    14 layers ResNet & 0.13 & 0.82 \\
    20 layers ResNet & 0.41 & 0.68 \\
    26 layers ResNet & 0.07 & 0.85 \\
    AttentiveFP-based GNN & 0.52 & 0.61 \\
    YZS-Model & \textbf{0.59} & \textbf{0.57} \\
    \hline
  \end{tabular*}
\end{table}

Furthermore, our research also utilized the dataset by Llinas et al. (2020) for additional model evaluation. Importantly, although the YZS-Model did not perform as expected on the test set1, it showed outstanding results on set2, with an RMSE of 1.05. This significantly outperforms the baseline model, AttentiveFP, which scored 1.28, and the SolTransNet model, which scored 1.24, as shown in Table \ref{tbl:Linans}. These results substantiate the model's effectiveness and potential for application.

\begin{table}[htbp]
  \centering
  \caption{SolTransNet Test Dataset Results Comparison}
  \label{tbl:Linans}
  \begin{tabular}{
    l
    S[table-format=1.2] 
    S[table-format=1.2]
    }
    \toprule
    & \multicolumn{2}{c}{RMSE} \\
    \cmidrule{2-3} 
    {Test set model} & {Llinas2020 set1} & {Llinas2020 set2} \\
    \midrule
    SolTransNet & 0.95 & 1.24 \\
    AttentiveFP-based GNN & 0.92 & 1.28 \\
    YZS-Model & \textbf{0.98} & \textbf{1.05} \\
    \bottomrule
  \end{tabular}
\end{table}

\begin{figure}[h]
  \centering
  \includegraphics[width=0.6\linewidth]{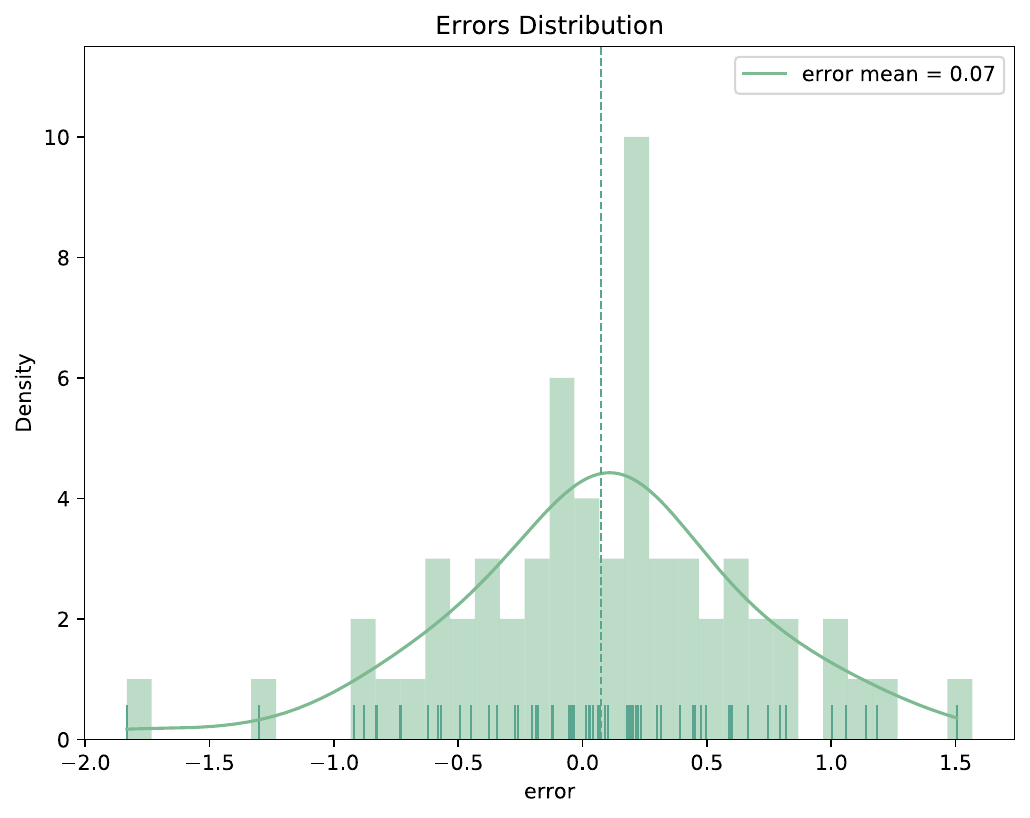}
  \caption{Error probability distribution of the YZS-Model on the test dataset.}
  \label{fig:fig4}
\end{figure}

Error analysis is depicted in Figure \ref{fig:fig4}, which shows the error probability distribution. This distribution plots the difference between the predicted logarithmic solubility values ($\hat y$) and the actual logarithmic solubility values ($y$) of the test molecules. With an error mean of 0.07, the figure highlights the average error level across the entire test set, visually demonstrating the distribution of errors within various ranges, with the majority of errors concentrated around the mean and larger errors being less frequent.

\subsection{Interpretative Analysis of YZS-Model}

In our model interpretation, we utilized the random feature zeroing technique to explore how atomic features influence the prediction biases for chemical molecules. Given that we employ one-hot encoding for the 66 elements, we chose to separately rank the atomic differences from other features. This separation enables a more accurate evaluation of each component's individual contribution to the prediction deviations.

\begin{figure}[h]
  \centering
  \includegraphics[width=0.9\linewidth]{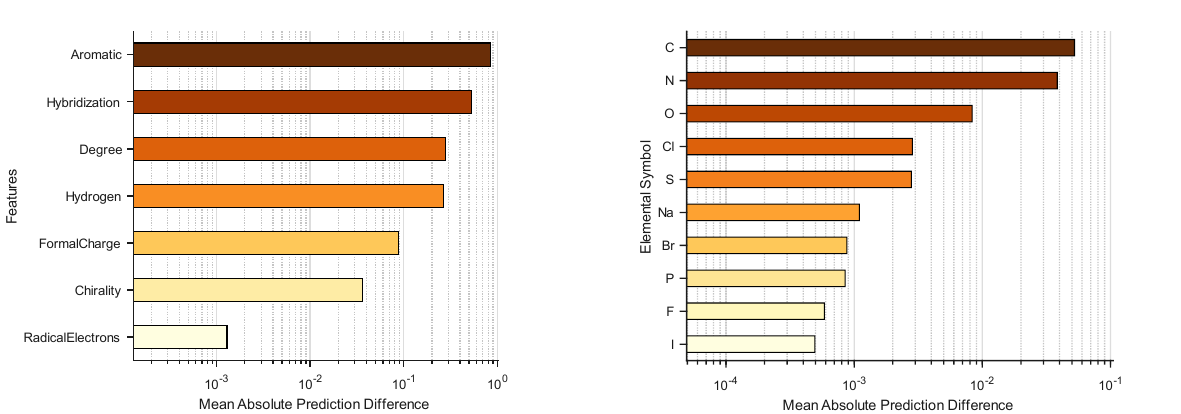}
  \caption{Ranking of Feature Importance using the Random Zeroing Method.}
  \label{fig:fig5}
\end{figure}

For the calculation of differences via random zeroing, we used an averaging approach—taking the average of a feature and its sub-features as the representative value for the entire feature. This method aims to consolidate the data from each sub-feature to more fully demonstrate the collective role of each feature and the one-hot encoded atoms in the model's prediction performance. The effects of this method, both in terms of quantification and qualitative analysis, are clearly illustrated in Figure \ref{fig:fig5} and detailed in the subsequent text.

In our research, we applied the technique of random feature zeroing to investigate the specific impacts of atomic features on prediction deviations for chemical molecules. Besides the element type, four features notably influenced predictions: aromaticity, hybridization type, the number of other atoms directly connected, and the number of directly connected hydrogens. Lesser influences were observed from features such as formal charge and stereoisomerism. Consequently, our primary analysis focused on these four features, with their importance indices detailed below:
\begin{table}[h]
\small
\centering
  \renewcommand{\arraystretch}{1.4}
  \caption{\ Mean Absolute Prediction Difference Index of Features}
  \label{tbl:index}
  \begin{tabular*}{0.25\textwidth}{@{\extracolsep{\fill}}ll}
    \hline
    Feature & MAPD \\
    \hline
    Aromatic & 0.8563\\
    Hybridization & 0.5253\\
    Degree & 0.2781\\
    Hydrogen & 0.2636\\
    \hline
  \end{tabular*}
\end{table}

\begin{figure}[h]
  \centering
  \includegraphics[width=0.3\linewidth]{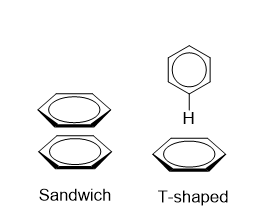}
  \caption{Aromatic rings' influence on molecular forces and electronic structures.}
  \label{fig:fig6}
\end{figure}

\begin{figure}[h]
  \centering
  \includegraphics[width=0.3\linewidth]{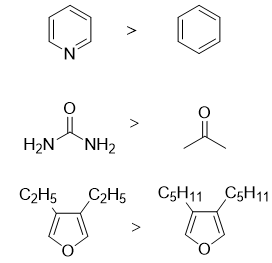}
  \caption{Solubility comparison among similar organic substances.}
  \label{fig:fig7}
\end{figure}

\begin{figure}[h]
  \centering
  \includegraphics[width=0.3\linewidth]{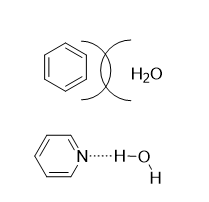}
  \caption{Diagram of benzene and pyridine interactions with water.}
  \label{fig:fig8}
\end{figure}

Aromaticity proved to have a significant impact on the model's solubility predictions, which was unexpected. This influence might stem from how aromatic rings alter molecular interactions and electronic structures, as depicted in Figure \ref{fig:fig6}\cite{frederick1990structural, headen2010structure}. 
Traditional solubility theories suggest that factors like polarity magnitude, number and strength of hydrogen bonds, and the balance of hydrophilic and hydrophobic groups primarily affect solubility (illustrated in Figure \ref{fig:fig7})\cite{martin1985dependence}. 
From these theories, we hypothesize that aromaticity's effect could be explained by the presence of benzene rings and other aromatic structures which might indicate lower polarity and thus, reduced solubility\cite{ruelle1998hydrophobic}. These rings also introduce significant hydrophobicity, whereas heteroaromatic rings, generally polar, can increase solubility, like pyridine which acts as a hydrogen bond donor (shown in Figure \ref{fig:fig8}).
This suggests that aromaticity's influence on solubility is complex, shaped by multiple overlapping factors. The requirement for specific hybridization (sp$^2$) on aromatic ring atoms further suggests a substantial role for hybridization in the predictions. The number of connected atoms, while related to atom type and hybridization, is less directly connected to aromaticity, possibly due to the diversity of atom types like C, N, O, etc. The number of connected hydrogens might relate to polarity and hydrophobic or hydrophilic nature, whereas connections to other atoms could complexly affect these properties, directly impacting solubility. Further experimental design is needed to explore these hypotheses.

\begin{figure}[h]
  \centering
  \includegraphics[width=1\linewidth]{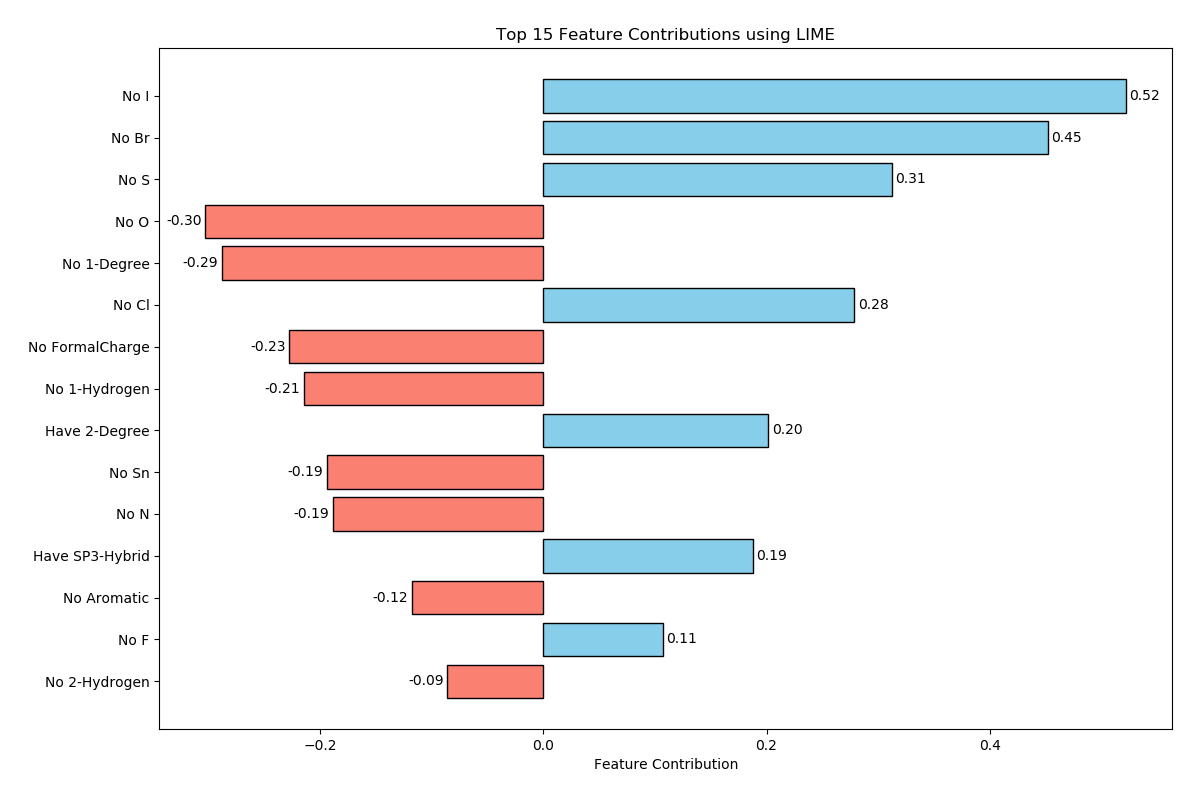}
  \caption{Ranking of individual features obtained via LIME, with red indicating negative influence and blue indicating positive influence.}
  \label{fig:fig11}
\end{figure}

Besides the explanations provided earlier, we also utilized the LIME technique to rank individual features by their importance and displayed the top 15 features as shown below Figure \ref{fig:fig11}. The sign on the x-axis represents whether the effect of the feature on solubility is positive (blue) or negative, where positive values indicate an increase in solubility, and negative values denote a decrease. The magnitude of the effect on solubility is depicted by the absolute value on the x-axis, with larger values representing a more significant impact. The y-axis illustrates the conditions under which each feature exerts its effect. For example, a positive score for "No Br" indicates that the absence of Br in an organic compound enhances solubility.

Figure \ref{fig:fig11} clearly shows that LIME identifies halogen atoms (F, Cl, Br, I, etc.) as having a substantial effect on molecular solubility. The lack of these atoms significantly increases solubility. Moreover, other factors with notable influence on the model's solubility predictions include the presence of S, N, and O, the degree of atom connectivity, the presence of formal charges, and the number of hydrogen atoms attached to an atom, along with hybridization and aromatic properties. These properties are categorized as follows: atom types, molecular complexity, and structural characteristics.

Atom Types and Their Impact on Solubility: In general, organic compounds with lower polarity exhibit lower solubility. LIME's results demonstrate that atom types significantly affect solubility. Halogen atoms reduce intermolecular polar interactions, decreasing solubility. In contrast, elements like S, N, and O form polar functional groups that increase molecular polarity and thus solubility. For example: Hydroxyl (-OH): Forms hydrogen bonds with water, significantly enhancing solubility. Amino (-NH$_2$) or Amide (-CONH$_2$): These functional groups are highly electronegative, improving solubility. Non-polar functional groups like thioether: Sulfur reduces molecular polarity, leading to lower solubility\cite{hou2004adme}.

Molecular Complexity and Connectivity: Formal Charge: Formal charge refers to the unequal sharing of electrons between atoms in some covalent bonds, resulting in some atoms within the molecule carrying partial positive charges while others bear partial negative charges. This distribution of charge increases the polarity of the molecule, thereby enhancing its solubility\cite{tsiper2001charge}. Degree of Connectivity: Connectivity refers to the number of bonds between atoms in a molecule. Higher connectivity indicates a more complex molecular structure, with more atomic connections. An increase in connectivity generally leads to increased molecular complexity as more atoms are interlinked to form larger molecular structures. More complex molecular structures may form more interactions within the solvent, which can affect the solubility of the molecule in the solvent. For instance, more complex structures may increase hydrophobicity, thus reducing solubility \cite{anderson1980solubility}.

Structural Properties: Aromaticity: Molecules with aromaticity typically possess a ring-shaped conjugated system, commonly found in aromatic compounds such as benzene rings. Functional groups with significant polarity, such as hydroxyl (-OH), carboxyl (-COOH), or amino (-NH$_2$) groups, can induce polar interactions on an aromatic ring, forming hydrogen bonds with water and enhancing the molecule’s interaction with water, thereby improving solubility\cite{bounds1981polarization}. Hybridization (sp$^2$ vs. sp$^3$): sp$^3$ hybridization tends to form more complex, non-planar structures that may increase intermolecular voids, providing more opportunities for polar interactions and thus improving solubility \cite{el2018linking}.

\begin{figure}[h]
  \centering
  \includegraphics[width=0.7\linewidth]{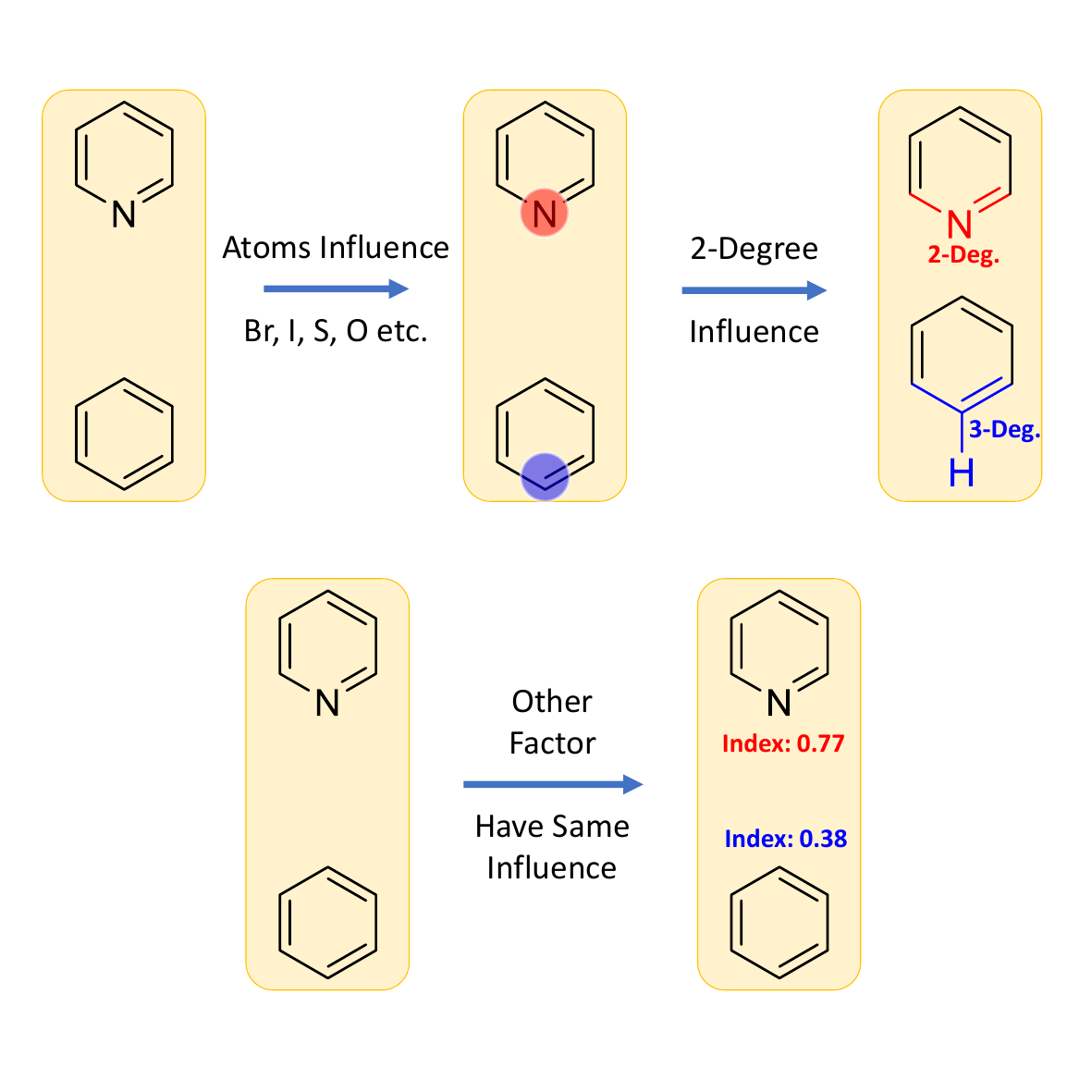}
  \caption{Qualitative solubility comparison between benzene and pyridine based on rankings provided by LIME.}
  \label{fig:fig12}
\end{figure}

To validate LIME's importance index, we predicted the solubility of pyridine and benzene. This involved calculating the top 15 solubility indices for these compounds to qualitatively assess solubility. As illustrated in Figure \ref{fig:fig12}, the YZS-model compared the solubility of benzene and pyridine: Impact of Atom Types: Benzene lacks the N element found in pyridine, leading to a solubility index 0.19 lower according to LIME. Degree of Connectivity: Pyridine's N atom is connected to two atoms, whereas benzene's corresponding position connects to three, resulting in a solubility index 0.20 lower for benzene. Other Factors: The compounds are otherwise identical, so no further analysis is needed. Ultimately, pyridine's solubility index is 0.39 higher than benzene's, consistent with actual solubility observations.

\subsection{Ablation experiment for YZS-Model}

\subsubsection{Evaluation of Training Data Dependency}

\begin{figure}[h]
  \centering
  \begin{minipage}{0.5\textwidth}
      \includegraphics[width=1\linewidth]{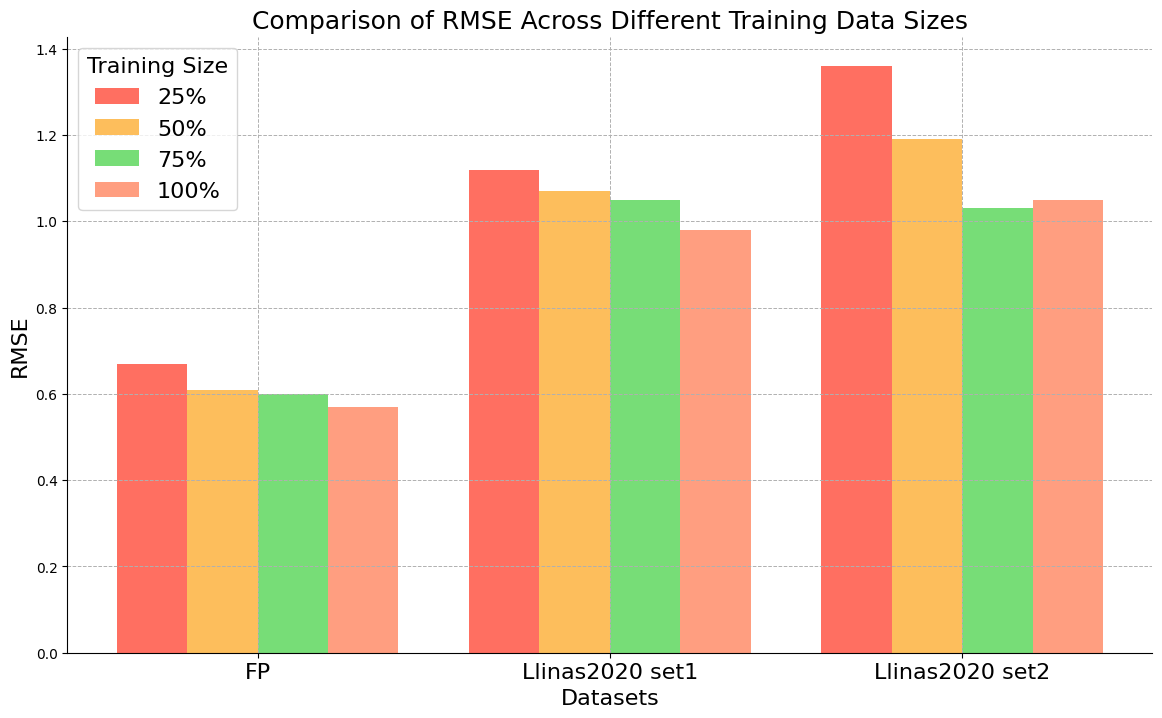}
      \caption{Performance results of the YZS-Model under different training set sizes in ablation studies.}
      \label{fig:exp1}
  \end{minipage}%
  \begin{minipage}{0.5\textwidth}
      Figure \ref{fig:exp1}. illustrates the impact of training data size on the model's RMSE across different datasets. In the ablation experiments, the training data was divided into subsets of 25\%, 50\%, 75\%, and 100\% of the original data size. The results indicate that using only 25\% of the training data leads to poor performance. However, increasing the training data size to 50\% significantly improves the predictive accuracy. Further increasing the data size to 75\% and 100\% continues to enhance the prediction results, although the performance gains beyond 75\% are marginal. This demonstrates that the model is highly dependent on the amount of training data, achieving baseline performance with just 50\% of the original training data.
  \end{minipage}
\end{figure}

To further clarify the sources of the model's interpretability performance, we performed a series of ablation experiments on the YZS-Model. The core principle of ablation experiments is to systematically control individual conditions or parameters and observe the changes in outcomes, thus revealing the contribution of different parts of the model to its overall performance. Firstly, to assess the model's dependence on the training set, we randomly shuffled the training dataset and then sequentially divided it into 25\%, 50\%, and 75\% of its original volume. We trained the model using these subsets, and the results are illustrated in Figure \ref{fig:exp1}.

As depicted in Figure 10, the model's performance is unsatisfactory when only 25\% of the original dataset is used. However, when the training set size is increased to 50\%, there is a notable improvement in prediction accuracy. Further increasing the training set to 75\% and 100\% continues to enhance the prediction results, although the performance gains become less significant after 75\%. This suggests that our model has a robust dependence on training data, achieving baseline performance with just 50\% of the original training data.

\subsubsection{Contribution Evaluation of YZS-Model Modules}

\begin{figure}[h]
  \centering
  \includegraphics[width=1\linewidth]{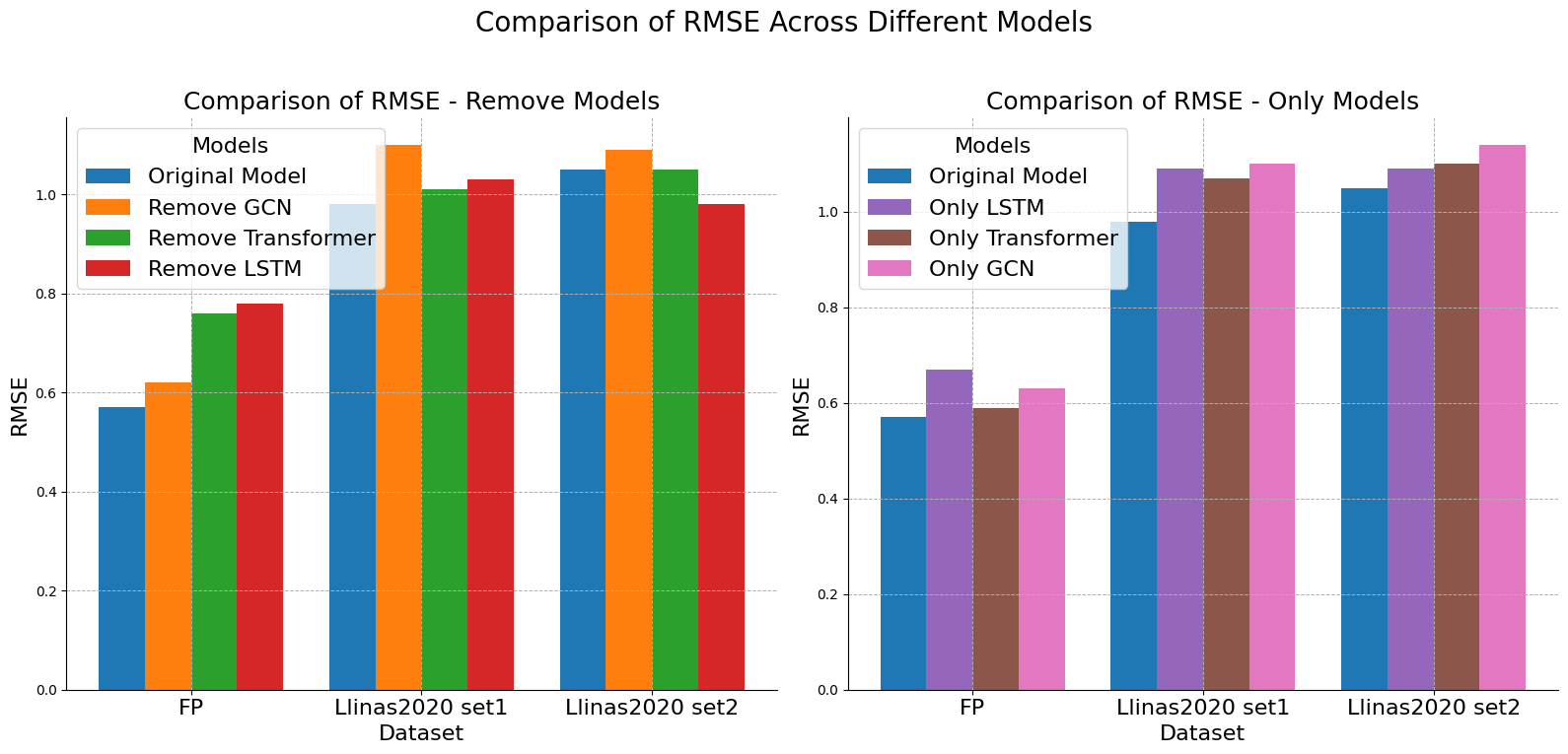}
  \caption{Performance results of the YZS-Model under different module combinations in ablation studies. The figure compares the RMSE of different models across various datasets. In the core component removal experiments shown on the left, the GCN, Transformer, and LSTM modules were individually removed and compared with the original model. The results indicate that removing the Transformer and LSTM modules has a significant impact on model performance, while the removal of the GCN module has a smaller effect. On the right, the single module experiments demonstrate that models using only LSTM, Transformer, or GCN individually perform worse than the original model, further confirming the importance of the synergistic effect of these modules in enhancing model performance.}
  \label{fig:exp2}
\end{figure}

To further explore the contributions of different modules in the YZS-Model during prediction, we conducted combination module ablation experiments. Specifically, we combined three essential submodules in various ways and performed ablation experiments on these combinations. This method allows us to determine the contribution of each module to the prediction results, leading to a deeper understanding of the model. The experimental results are presented in Figure \ref{fig:exp2}.

The figure compares the RMSE of different models across various datasets. In the core component removal experiments shown on the left, the GCN, Transformer, and LSTM modules were individually removed and compared with the original model. The results indicate that removing the Transformer and LSTM modules has a significant impact on model performance, while the removal of the GCN module has a smaller effect. On the right, the single module experiments demonstrate that models using only LSTM, Transformer, or GCN individually perform worse than the original model, further confirming the importance of the synergistic effect of these modules in enhancing model performance.

Initially, by examining the performance of the original model, we observe that the complete model achieves an RMSE of 0.57, significantly outperforming other configurations, indicating that the model's integrity is critical for prediction performance. Next, we analyzed the ablation effects of individual modules. By separately removing the GCN, Transformer, and LSTM modules, we found that the exclusion of the GCN module had a minor impact on prediction performance, whereas the removal of the Transformer and LSTM modules had a more substantial effect. This suggests that while the GCN is involved in feature extraction, it has a lesser impact on prediction performance compared to the more crucial roles of the Transformer in capturing complex feature relationships and the LSTM in uncovering deep sequential dependencies within molecular sequences. Further analysis of dual-module ablation experiments indicates that the model performs best when both the GCN and LSTM are excluded, highlighting the significant positive influence of the Transformer module on prediction results. This conclusion is corroborated by other dual-module ablation experiments: prediction performance significantly deteriorates when the Transformer is absent but the LSTM or GCN are present. Therefore, we can conclude that the Transformer is the key predictive module in the YZS-Model.

\subsubsection{Detailed Evaluation of Transformer Module}

To further investigate the contributions of the internal components of the Transformer model to predictive accuracy, and to elucidate its internal mechanisms and the importance of its components, we conducted two types of ablation experiments. The primary objective was to identify and quantify the impact of key components—Multi-Head Self-Attention (MSA) and Feed-Forward Neural Network (FFN)—as well as the number of layers on overall model performance. We employed the following methods.

Core Component Ablation: In the first set of experiments, we replaced core components of the Transformer—MSA and FFN—with identity mappings. Specifically, the MSA mechanism was substituted with an identity mapping, allowing inputs to bypass the self-attention computation and directly pass to the next layer. Similarly, the FFN was replaced with an identity mapping, permitting inputs to proceed to the subsequent layer without undergoing the feed-forward transformation. These modifications aimed to assess the importance of MSA and FFN in learning complex features and enhancing predictive accuracy.

Layer Ablation: In the second set of experiments, we varied the number of Transformer layers to observe the effects of different model depths on prediction performance. We tested configurations with reduced layers, specifically architectures with 2, 4, and 6 layers, and configurations with increased layers, specifically architectures with 8, 10, and 12 layers. The goal was to understand how the number of layers influences the model's ability to capture decisive factors in the data and balance between model complexity and overfitting.

\begin{figure}[h]
  \centering
  \includegraphics[width=1\linewidth]{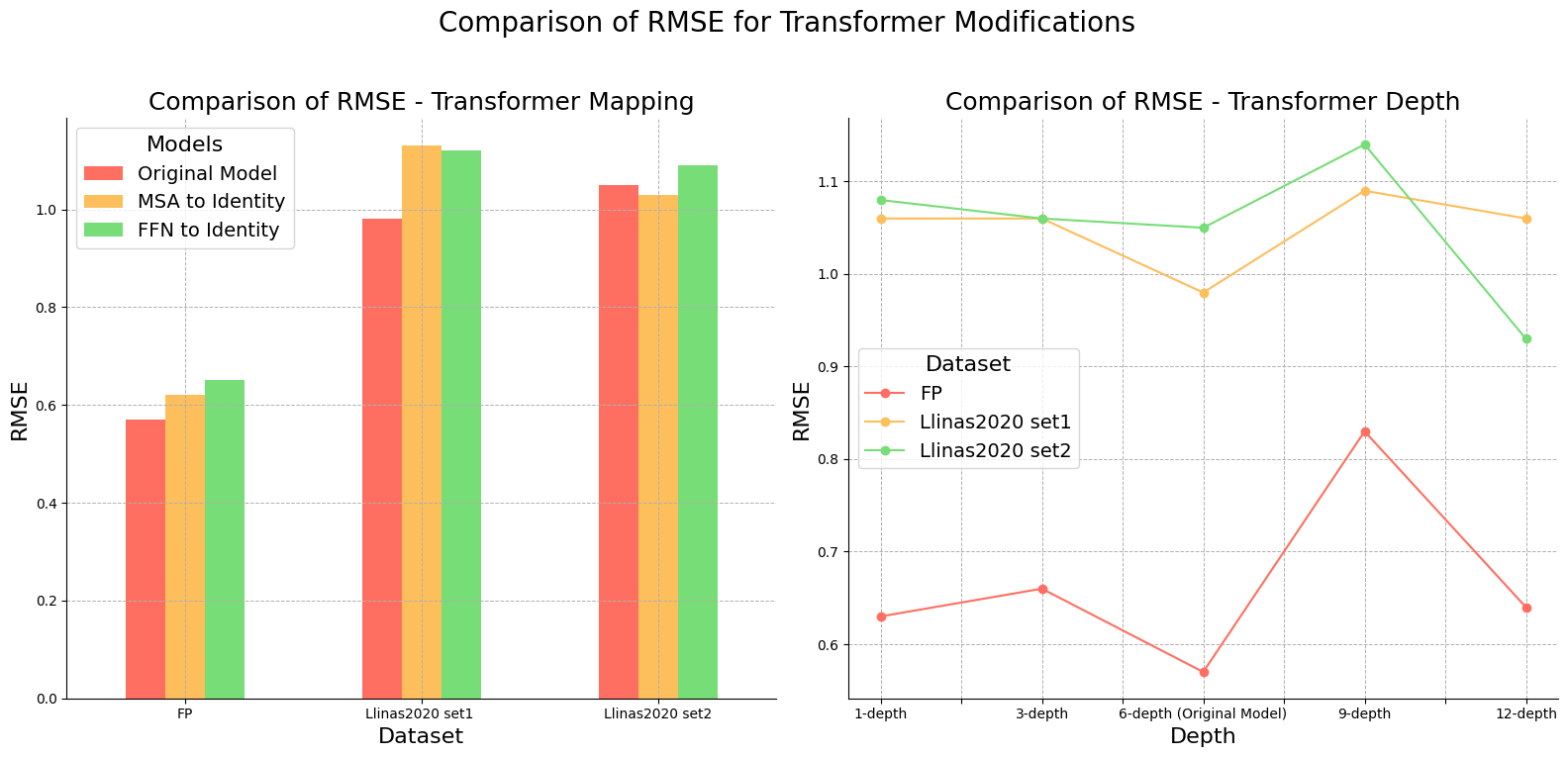}
  \caption{Performance results of the YZS-Model under ablation of specific internal modules of the Transformer. Figure illustrates the impact of replacing core components and adjusting the number of layers on the model's RMSE across different datasets. In the core component ablation experiments, the Multi-Head Self-Attention (MSA) and Feed-Forward Neural Network (FFN) were replaced with identity mappings. The results indicate that both MSA and FFN are crucial to the Transformer's attention mechanism. In the layer ablation experiments, Transformer architectures with 2, 4, 6, 8, 10, and 12 layers were tested. It was found that the 6-layer Transformer consistently performed the best across multiple datasets.}
  \label{fig:exp3}
\end{figure}

As figure \ref{fig:exp3}, in the core component ablation experiments, replacing MSA or FFN with identity mappings resulted in a significant increase in RMSE, underscoring the crucial role of these components in the Transformer's attention mechanism. Specifically, replacing MSA led to a marked increase in RMSE, highlighting its essential function in capturing intricate relationships within the input data. Similarly, replacing FFN resulted in a significant RMSE increase, emphasizing its vital role in processing and transforming feature representations. In the layer ablation experiments, RMSE exhibited a trend of initially decreasing and then increasing with the number of layers. Models with 2 and 4 layers had high RMSE, suggesting underfitting due to inadequate capacity to capture key features. The 6-layer Transformer achieved the lowest RMSE, indicating an optimal balance where the model sufficiently captured features without overfitting. However, models with 8, 10, and 12 layers showed increased RMSE, suggesting overfitting due to excessive complexity. Notably, the 12-layer Transformer performed best on the set2 dataset but exhibited instability across other test datasets, likely due to randomness in the training process. Overall, the results identified the 6-layer Transformer as the most suitable architecture for this task, offering an optimal trade-off between model complexity and generalization ability.

These experimental results provide valuable insights into the contributions of various components within the YZS-Model to prediction tasks, forming a basis for further model optimization. The ablation experiments not only confirmed the critical role of the Transformer module but also highlighted the specific effects of its internal components and layer number on model performance, guiding future research and applications.

\subsection{Discussion}

This research highlights the critical role of considering specific molecular structural features during the design and training stages of the model, as well as the necessity for specifically optimizing these features to improve model performance. Future initiatives will aim to enhance the model's adaptability and precision in predicting rare molecular structures, ensuring superior results across a more extensive dataset range.

Although the YZS-Model exhibited excellent performance across most testing datasets, its performance on the Llinas2020 set1 did not reach the anticipated levels and failed to exceed existing benchmark models. This observation suggests that the Llinas2020 set1 dataset might include unique molecular structures that are seldom present in the training set, which hinders the model's ability to learn their essential features fully. Moreover, despite the theoretical capability of the GCN and Transformer mechanism to effectively identify complex structures and long-range dependencies, the model exhibits certain limitations when dealing with very rare molecular structures.

\begin{figure}[h]
  \centering
  \includegraphics[width=0.7\linewidth]{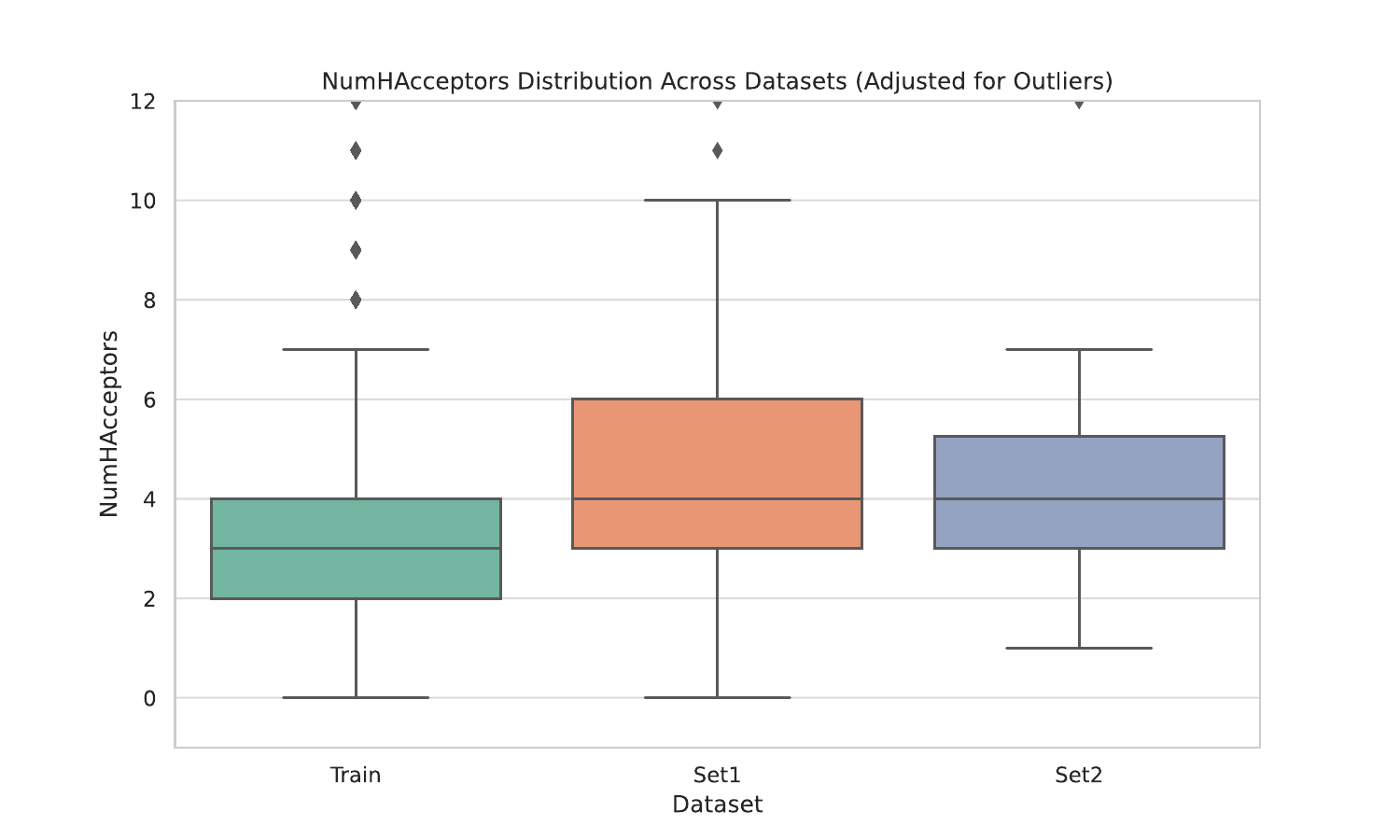}
  \caption{Potential differences in the number of hydrogen bond acceptors in the Set1 dataset.}
  \label{fig:fig9}
\end{figure}

Further insights are provided by an in-depth analysis of the NumHAcceptors distribution. The Llinas2020 set1 contains more molecules with a high number of hydrogen bond acceptors compared to the training and other test sets, as depicted in Figure \ref{fig:fig9}, potentially explaining the model's poor performance on this particular dataset.

Furthermore, our analysis suggests that the model's performance discrepancies across different datasets could also be attributed to the number of functional groups in the molecules. Molecules with more functional groups are inherently more complex, increasing the likelihood of feature loss during the feature aggregation phase. Notably, the data distribution of the Llinas2020 set2 test dataset shows high similarity to our training dataset sourced from Cui, Q. However, a significant difference exists with Llinas2020 set1, and we hypothesize that the unusual complexity of Set1 could be a critical factor affecting prediction outcomes.

\begin{figure}[!h]
  \centering
  \includegraphics[width=0.7\linewidth]{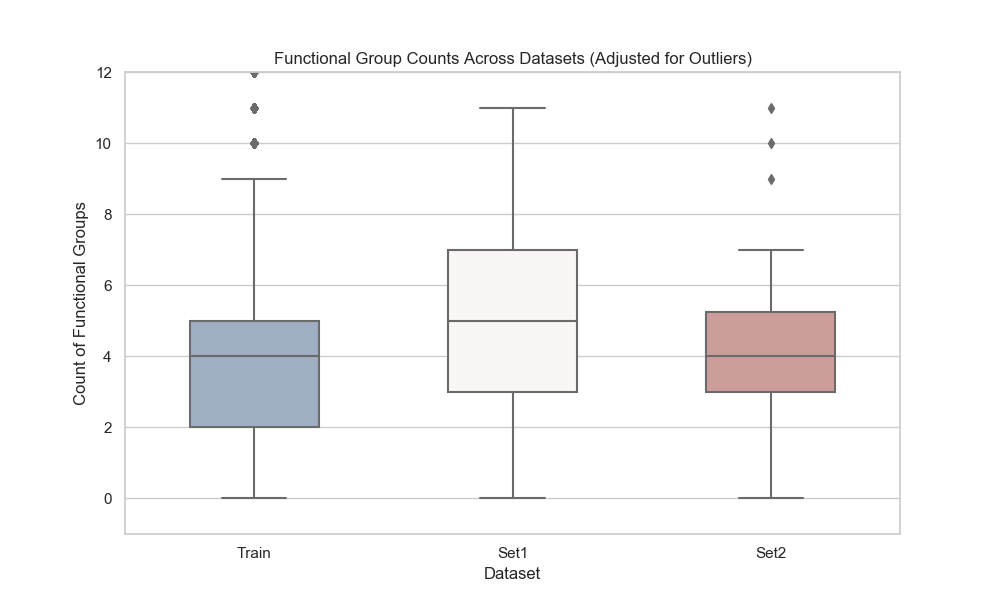}
  \caption{Potential differences in the number of functional groups (complexity) in the Set1 dataset.}
  \label{fig:fig10}
\end{figure}

Our study reveals further potential limitations of the model, especially when processing molecules with specific and rare structural traits or complex features, where there is room for improvement in prediction accuracy. Future research could enhance model generalization by incorporating a more varied range of datasets or by implementing more sophisticated feature extraction and learning methods to refine model performance.

Significantly, our model shows extensive potential for application in drug design and development. Precise solubility predictions in high-throughput screening can assist researchers in more effectively identifying and optimizing potential drug candidates, thus speeding up the drug development pipeline and lowering costs. Additionally, the accuracy and generalization capabilities of the YZS-Model offer novel insights into the intricate connections between molecular structures and their physicochemical properties.

\section{Conclusion}
In this study, we developed and validated the YZS-Model, an innovative deep learning framework based on Graph Convolutional Networks (GCNs), Transformer architecture, and Long Short-Term Memory networks (LSTMs), designed for the precise prediction of organic drug solubility. By integrating these three advanced deep learning technologies, the YZS-Model can deeply analyze the spatial structure and sequence dynamics of drug molecules, resulting in significant performance improvements in the prediction of organic drug solubility.

Our model has demonstrated superior performance across different test sets, particularly excelling in solubility predictions for anticancer compounds, where it outperformed existing benchmark models in terms of accuracy and generalization capability. Additionally, through a series of detailed feature importance analyses, we have identified key factors affecting drug molecule solubility, providing valuable insights for drug design and optimization.

However, we also observed that the model's accuracy could be further improved when dealing with molecules with specific and rare structural features. To address this challenge, future research will consider further expanding the model by exploring other machine learning strategies, including semi-supervised and unsupervised learning methods, to utilize unlabeled data for model training, thereby enhancing the model's generalization ability to unknown datasets. We also plan to introduce multi-scale attention mechanisms to more precisely capture the complex interactions within molecules, improving prediction accuracy. Additionally, ensemble learning strategies will be employed to further enhance prediction performance.

In summary, the YZS-Model not only demonstrates strong potential in predicting the solubility of organic drugs but also paves new avenues for the application of deep learning technologies in drug discovery and development. The application of deep learning technologies is expected to assist in preclinical drug screening and optimize the chemical properties of candidate drugs to increase their bioavailability. Moreover, in the field of personalized medicine, this model holds promise for integrating individual genetic backgrounds to predict drug responses, thereby informing the development of personalized treatment strategies. We anticipate that future research will further explore and expand these technologies, making greater contributions to human health.

Despite our achievements, it is essential to recognize that drug development is a complex and dynamic process, involving factors far beyond the explanatory power of a single model. Therefore, we encourage ongoing exploration and improvement by future researchers to fully leverage the vast potential of deep learning in drug development.

\bibliographystyle{unsrt}  
\bibliography{references}

\end{document}